\pdfoutput=1
\documentclass[11pt]{article}
\usepackage{amsmath}
\usepackage{graphicx}
\usepackage{ACL2023}

\usepackage{times}
\usepackage{hyperref}
\usepackage{latexsym}
\usepackage{tabularx}
\usepackage{booktabs}
\usepackage{array}
\usepackage[T1]{fontenc}
\usepackage[utf8]{inputenc}

\usepackage{microtype}
\usepackage{amssymb}
\usepackage{inconsolata}

\title{A Survey of Calibration Process for Black-Box LLMs}

\author{Liangru Xie$^{1}\thanks{\hspace{4pt}Work done during an internship at Amazon.}$, Hui Liu$^1$, Jingying Zeng$^1$, Xianfeng Tang$^1$,\\
{\bf Yan Han$^1$, Chen Luo$^1$, Jing Huang$^1$, Zhen Li$^1$, Suhang Wang$^2$, Qi He$^1$}\\
  \textsuperscript{1}Amazon \ \ \ \ \ \ \textsuperscript{2}Penn State University \\
  \texttt{xielr@ieee.org, \{liunhu, zejingyi, xianft\}@amazon.com, szw494@psu.edu} \\
  }

\begin{document}
\maketitle
% \begin{flushleft}
% \textsuperscript{*}Work done during an internship at Amazon.
% \end{flushleft}
\begin{abstract}
Large Language Models (LLMs) demonstrate remarkable performance in semantic understanding and generation, yet accurately assessing their output reliability remains a significant challenge. While numerous studies have explored calibration techniques, they primarily focus on \textit{White-Box LLMs} with accessible parameters. \textit{Black-Box LLMs}, despite their superior performance, pose heightened requirements for calibration techniques due to their API-only interaction constraints. Although recent researches have achieved breakthroughs in black-box LLMs calibration, a systematic survey of these methodologies is still lacking.  To bridge this gap, we presents the first comprehensive survey on calibration techniques for black-box LLMs. We first define the \textit{Calibration Process} of LLMs as comprising two interrelated key steps: \textit{Confidence Estimation} and \textit{Calibration}. Second, we conduct a systematic review of applicable methods within black-box settings, and provide insights on the unique challenges and connections in implementing these key steps. Furthermore, we explore typical applications of Calibration Process in black-box LLMs and outline promising future research directions, providing new perspectives for enhancing reliability and human-machine alignment. This is our GitHub link: \href{https://github.com/LiangruXie/Calibration-Process-in-Black-Box-LLMs}{https://github.com/LiangruXie/Calibration-Process-in-Black-Box-LLMs}
\end{abstract}

\section{Introduction}
Large Language Models (LLMs) have demonstrated exceptional generalization abilities and contextual understanding across various application scenarios \cite{chang2024survey}, particularly in open-domain question-answering tasks \cite{srivastava2024towards, wang2023aligning}. However, when faced with ambiguous prompts or insufficient domain knowledge \cite{guu2020retrieval, feng2024don, guo2017calibration, ji2023towards}, LLMs often produce hallucinations \cite{kryscinski2019evaluating, liu2024exploring, manakul2023selfcheckgpt, perkovic2024hallucinations} or exhibit overconfidence in their responses \cite{zhao2021calibrate, yang2024can}. While traditional solutions such as Reinforcement Learning with Human Feedback (RLHF) \cite{wang2024offline} and parameter-level interventions (e.g., model weight modification, adapter tuning) have been proposed for white-box LLMs where model checkpoints and architectures are accessible (like Llama \cite{touvron2023llama}, ChatGLM \cite{glm2024chatglm}, and Vicuna \cite{chiang2023vicuna}), these parameter-dependent approaches prove ineffective for black-box LLMs such as GPT \cite{achiam2023gpt}, Claude \cite{caruccio2024claude}, and Gemini \cite{team2023gemini}, which only allow input-output interactions through API calls. Given these limitations and the high computational costs of traditional methods, Confidence Estimation and Calibration Methods have emerged as promising alternatives \cite{ni2024llms, mielke2022reducing}, offering cost-effective solutions applicable to both white-box and black-box models while maintaining scalability to large-scale deployments.
\begin{figure*}
  \centering
    % \fbox{\rule{0pt}{2in} \rule{.9\linewidth}{0pt}}
  \includegraphics[width= \linewidth]{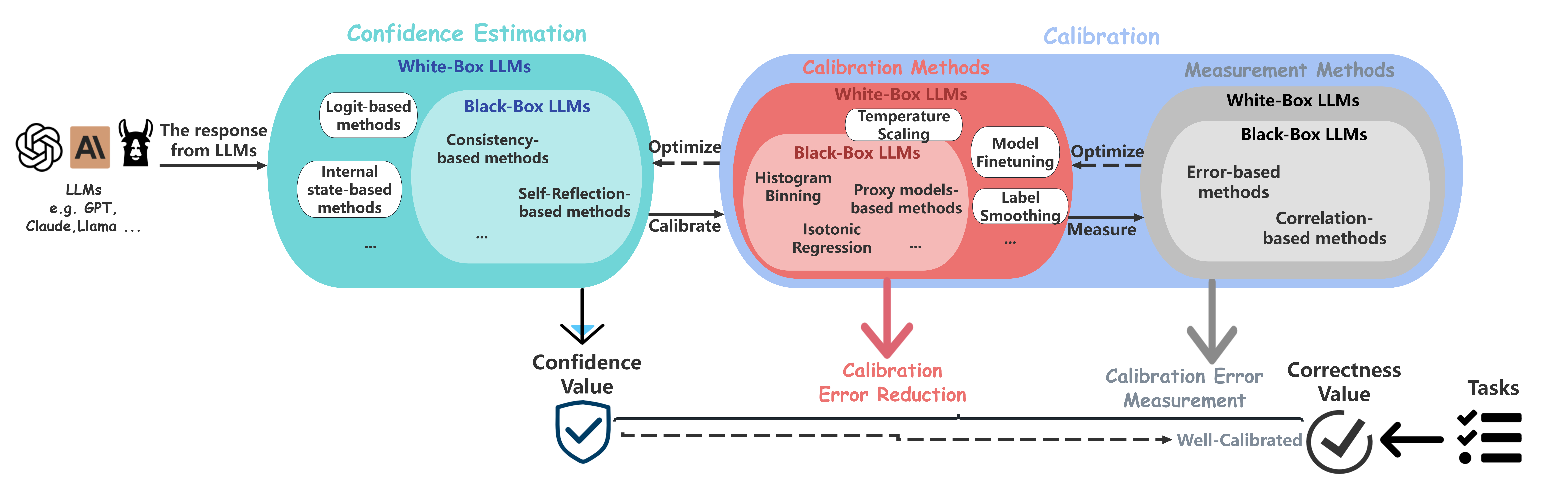}
  \caption{Calibration Process. As shown in this figure, LLM responses first go through the Confidence Estimation module to obtain confidence values, while correctness values are determined based on the specific task. In the Calibration module, Calibration Methods reduce the calibration error by minimizing the gap between confidence and correctness values, followed by Measurement Methods to assess the actual calibration error. The primary objective of the entire calibration process is to obtain well-calibrated confidence values that accurately reflect the response quality. Additionally, this figure demonstrates that methods used in black-box LLMs are fully incorporated into each module of white-box LLMs, while the application of white-box methods in black-box LLMs remains limited.}
  \label{fig:img0}
\end{figure*}

In recent years, Confidence Estimation and Calibration have frequently been discussed together, as the estimation of confidence is often influenced by the uncertainty in the model or data, and calibration methods help the model recognize its own knowledge limitations \cite{jiang2021can, shrivastava2023llamas}. Calibration allows LLMs to adjust their confidence to more accurately reflect the quality of their outputs \cite{kuhn2023semantic, duan2023shifting}. For example, in the context of diagnosing rare diseases, LLMs may estimate a 95\% confidence score to an incorrect response, while the accurate confidence, due to a lack of domain knowledge, should be closer to 40\%. Calibration can identify this discrepancy and adjust the model's confidence to more accurately reflect the quality of the response, preventing overconfidence in generated responses \cite{ren2023self, geng2024survey, tian2023just}. This process of Confidence Estimation, and Calibration to achieve \textit{well-calibrated} confidence is referred to as the \textit{Calibration Process} in this survey.

% \hui{Maybe make some order adjustment here. In the following paragraph, we are discussing BB LLMs. Then in the next one, we discuss calibration, and then we go back to discuss BB LLMs again. Maybe we can start from calibration, and then go to BB LLMs.}
\begin{figure*}
  \centering
    % \fbox{\rule{0pt}{2in} \rule{.9\linewidth}{0pt}}
  \includegraphics[width= \linewidth]{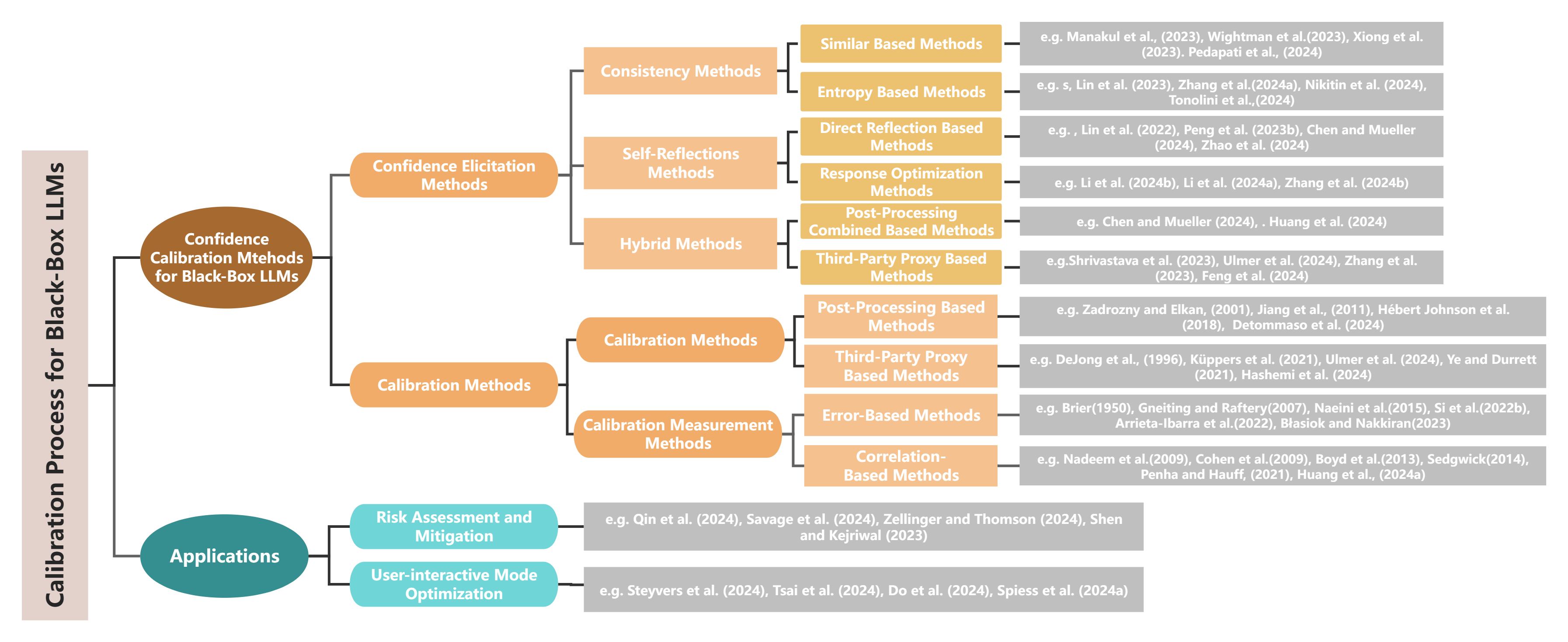}
  \caption{Organizational Structure of the Calibration Process for black-box LLMs.}
  \label{fig:img1}
\end{figure*}
A growing number of surveys have reviewed factors influencing LLMs Calibration Process. For instance, some studies have summarized how model parameters, training stages, or training data influence confidence and calibration \cite{zhu2023calibration}, while others investigate which confidence representations are more beneficial for calibration during the RLHF phase \cite{tian2023just}. Additionally, many studies have reviewed various confidence estimation and calibration methods on Language Models(LMs) and LLMs \cite{geng2024survey}. However, existing surveys predominantly focus on white-box LLMs, where Calibration Process methods typically encompass logit-based calibration \cite{bi2024decoding, huang2023look, kuhn2023semantic}, additional parameter layers \cite{liu2024litcab}, instruction tuning \cite{kapoor2024calibration}, and reinforcement learning techniques \cite{band2024linguistic}.

In contrast to white-box LLMs, research on the Calibration Process for black-box LLMs lacks systematic analysis despite active development. black-box LLMs have recently garnered significant attention due to their exceptional performance across various real-world applications, especially as benchmark judge models for evaluating LLMs output quality and overall performance \cite{zheng2023judging}. This role has been widely adopted by small and medium-sized enterprises for product development \cite{lin2023generating, li2024think, chen2024quantifying}. Given black-box LLMs’ function as critical judge models, enhancing their sensitivity to uncertainty, improving output reliability, and fostering user trust are essential. The Calibration Process has demonstrated substantial potential in meeting these requirements \cite{wagner2024black}, further driving interest in black-box LLMs Calibration Process research \cite{si2022prompting, ye2022unreliability, lin2023generating, li2024think, chen2024quantifying}.

Despite this growing interest, no comprehensive study, to the best of our knowledge, has been dedicated specifically to the Calibration Process for black-box LLMs. This survey aims to bridge this gap by providing a systematic review and analysis of current research progress in this field. Our investigation reveals that existing Calibration Process methods for black-box LLMs can be categorized into two main approaches: the introduction of third-party proxy models \cite{ulmer2024calibrating, shen2024thermometer} that partially transform black-box models into gray-box models, and post-processing techniques that operate solely on input-output data. The complexity of calibrating black-box LLMs is further exacerbated by the inherent challenge of inaccessible internal logic information \cite{10.1145/3236009, jiang2018trust}, highlighting the critical importance of this survey in advancing the field of Calibration Process.

Importantly, since methods for black-box settings are often applicable to white-box as well, studying the Calibration Process of black-box LLMs not only addresses unique challenges in black-box settings but also provides valuable insights for the calibration of white-box LLMs, as illustrated in Figure \ref{fig:img0}. This comprehensive study encompasses two key aspects in black-box settings: Confidence Estimation, and Calibration. While Calibration includes measurement methods for assessing calibration error, these measurement methods are highly interoperable across black-box and white-box LLMs and will not be extensively discussed. These methodologies will be thoroughly explored in Section \ref{cpm}, followed by their practical applications in Section \ref{app}. Section \ref{fd} concludes with a discussion of future research directions for black-box LLMs Calibration Process, with the overall structure of our survey presented in Figure \ref{fig:img1}.

\section{Calibration Process Methods of black-box LLMs}
\label{cpm}
As shown in Figure \ref{fig:img0}, the Calibration Process is a cyclical and consists of two main steps: Confidence Estimation and Calibration, where the Calibration step is further divided into calibration and measurement phases. This process aims to obtain well-calibrated confidence values that accurately reflect the quality of the response and align with correctness. This Calibration Process can be represented by Formula \ref{eq:well-callibration}:
\begin{equation}
V(\text{Correct} \mid \text{Confidence} = v) = v
\label{eq:well-callibration}
\end{equation}

In the Formula \ref{eq:well-callibration}, $V$ represents the response correctness value, while $v$ represents the confidence value. It is worth noting that while white-box and black-box LLMs share the same Calibration Process steps, many methods suitable for white-box models cannot be directly applied to black-box settings due to their inherent closed nature. Therefore, this section focuses on Confidence Estimation and Calibration methods applicable to black-box LLMs, with comprehensive coverage of measurement methods in the Calibration portion.

\begin{table*}[h]
\scriptsize
% \small
\begin{tabularx}{\linewidth}{m{2cm}<{\centering} m{6cm}<{\centering} m{3cm}<{\centering} m{2cm}<{\centering} m{1cm}<{\centering}}
	\hline
Study & Proposed Methods & Datasets & Model & Category\\ 
 
    \hline
\citet{xiong2023can} & Introduced a multivariate Confidence Estimation framework combining self-random, prompt-based, and adversarial sampling methods. A similarity-based aggregator integrates verbalized confidence and Pair-Rank, optimizing well-calibrated confidence across tasks. & SportUND \cite{xia2024sportqa}, StrategyQA \cite{geva2021did}, GSM8K \cite{cobbe2021training}, SVAMP \cite{patel2021nlp}, DateUnd \cite{ghazal2013bigbench}, ObjectCou \cite{wang2019learning}, Prf-Law \citet{hendrycks2020measuring}, Biz-Ethics \cite{hendrycks2020measuring} & GPT-3 175B \cite{brown2020language}, GPT-3.5-turbo \cite{ye2023comprehensive}, GPT-4 \cite{achiam2023gpt} & Consistency \\
    \hline
\citet{zhang2024luq} & Addressed Confidence Estimation in long-form texts by leveraging uncertainty. They measured the non-contradiction probability between sentences in response samples to estimate uncertainty. & FactScore dataset \cite{min2023factscore} & GPT-4 \cite{achiam2023gpt}, GPT-3.5 \cite{abdullah2022chatgpt}, Gemini 1.0 Pro \cite{team2023gemini} & Consistency  \\
    \hline
\citet{li2024confidence} & Proposed the 'If-or-Else' (IoE) prompting framework, where LLMs either retain or revise their answers based on confidence. Confidence is inferred from response consistency, with unchanged answers indicating higher confidence. & GSM8K \cite{cobbe2021training}, SVAMP \cite{patel2021nlp}, HotpotQA \cite{yang2018hotpotqa}, Sports \citet{xia2024sportqa}, LLC \cite{wei2022chain}, Domestic Robot \cite{cai2023benchlmm} & GPT-3.5-0613, GPT-3.5-1106 \cite{abdullah2022chatgpt}, GPT-4, GPT-4V \cite{achiam2023gpt} & Self-Reflection   \\
    \hline
\citet{zhao2024fact} & Explored the impact of prompt designs like Chain-of-Thought \cite{wei2022chain}, Tree-of-Thought \cite{yao2024tree}, and Self-Ask \cite{press2022measuring} on confidence, generating intermediate confidence values for final estimation. & StrategyQA \cite{geva2021did}, Web Questions \cite{bordes2014question}, Human Eval dataset \cite{wang2022self} & GPT-3 \cite{abdullah2022chatgpt} & Self-Reflection \\
\hline
\citet{huang2024calibrating} & Used ChatGPT-4 \cite{achiam2023gpt} to generate correctness and confidence score distribution curves, then applied similarity-based metrics to compute calibration errors and adjust for optimal confidence values. & ASQA \cite{stelmakh2022asqa}, ELI5 \cite{fan2019eli5}, QAMPARI \cite{amouyal2022qampari}, CNNDM \cite{nallapati2016abstractive}. & GPT-3.5-turbo \cite{ye2023comprehensive} & Hybrid Methods \\
\hline
\citet{becker2024cycles} & Generates multiple explanations from the LLM, reweights them using entailment probability for logical consistency, and computes a posterior distribution to derive a reliable confidence score. & CSQA \cite{talmor2018commonsenseqa}, TruthQA \cite{lin2021truthfulqa}, MedQA \cite{jin2021disease}, MMLU Law, MMLU Physics \cite{hendrycks2020measuring}. & GPT-3.5-turbo \cite{ye2023comprehensive}, GPT-4-turbo \cite{achiam2023gpt} & Hybrid Methods \\
\hline
\end{tabularx}
    \caption{Table of Latest Confidence Estimation Methods.}
    \label{tab:cem}
\end{table*}

\subsection{Confidence Estimation}
\label{sec:ccm}
For black-box LLMs, Confidence Estimation depends exclusively on input-output information, leading researchers to develop methods that extract reference information from model outputs through designed interactions and multiple queries. While some studies term this process ''Confidence Elicitation'' \cite{xiong2023can, huang2024calibrating, shrivastava2023llamas}, we maintain the term ''Confidence Estimation''.

Two main approaches exist: \textit{Consistency} and \textit{Self-Reflection}, both post-processing techniques that enable near-well-calibrated estimation without accessing model parameters. These methods can be employed individually or integrated into hybrid approaches, sometimes leveraging third-party proxy models. Furthermore, beyond the methods discussed in the following sections, additional latest methods are summarized in Table \ref{tab:cem}.

% In black-box LLMs, Confidence Estimation relies solely on input and output information. As a result, many studies incorporate various approaches to extract sufficient reference information from the LLM's output during the Confidence Estimation. For example, by designing specific interactions or multiple query processes to obtain various outputs, exploring the limits of the output space, and deriving relatively reliable confidence values. Some research refers to this process as ''Confidence Elicitation'' based on its characteristics \cite{xiong2023can, huang2024calibrating, shrivastava2023llamas}. To avoid confusion, we will continue using the term "Confidence Estimation" in our study. 

% We summarize two Confidence Estimation methods applicable to black-box LLMs: \textit{Consistency} and \textit{Self-Reflection}. Both methods are post-processing techniques that do not require any internal parameter information of the model, enabling near-well-calibrated Confidence Estimation for black-box LLMs from both external and internal perspectives. In this section, we will first introduce the  \textit{Consistency} and \textit{Self-Reflection} methods separately.  Subsequently, in the \textit{Hybrid Methods}, we will discuss methods that combine the strengths of both, as well as introduce some methods that involve third-party proxy models for Confidence Estimation. The latest methods is shown in Table \ref{tab:cem}.

\subsubsection{Consistency Methods}
Consistency methods enable effective estimation of black-box LLMs by capturing the relationships and variations among model responses. These methods fall into two categories: similarity-based and entropy-based approaches. While similarity measurements directly yield confidence scores, entropy calculations indicate uncertainty levels that inversely reflect confidence, where confidence measures the stability of model responses and uncertainty captures their potential variability.

In similarity-based methods, for instance, selfCheckGPT \cite{manakul2023selfcheckgpt} uses metrics like BERTScore \cite{zhang2019bertscore}, MQAG \cite{manakul2023mqag}, n-gram, NLI \cite{he2021debertav3}, or GPT-3 (text-davinci-003) prompts \cite{ye2023comprehensive} to evaluate the similarity between response samples of the same query and thus derive the confidence values. \citet{wightman2023strength} proposed a framework to assess the similarity between responses generated by different prompts containing the same question, introducing a prompt generation mechanism that calculates confidence values based on the frequency of responses across multiple prompts. They demonstrated that more prompts improve confidence calibration. \citet{pedapati2024large} introduced six methods for expanding output variable space in black-box models, namely Stochastic Decoding, Split Response Consistency, Paraphrasing, Sentence Permutation, Entity Frequency Amplification, and Stop-word Removal. The lexical and semantic similarity features between responses generated by each method are then extracted and input into logistic regression models to map these features to reliable confidence values.

In entropy-based methods, \citet{lin2023generating} introduced a framework for uncertainty quantification using methods such as semantic set quantity, graph Laplacian eigenvalues, and degree matrix to measure uncertainty among response samples. \citet{nikitin2024kernel} proposed a more comprehensive approach, first calculating semantic similarity between response samples using a PSD semantic kernel, constructing a semantic graph with response samples as nodes and similarities as edges, and finally quantifying uncertainty using von-Neumann entropy. \citet{tonolini2024bayesian} propose a Bayesian Prompt Ensembles method that assigns weights to semantically equivalent prompts based on their inherent uncertainty, integrating them to compute a lower bound on model error and produces near-well-calibrated uncertainty estimates.

% \noindent\textbf{Advantages and Disadvantages of Consistency.}
% Consistency methods have been proven to be closely related to correctness, with higher consistency typically indicating greater accuracy. Moreover, consistency methods fall under external Confidence Elicitation techniques, which do not require altering the internal state of the model. Instead, they rely solely on the original query and response for calculations, making them applicable to a variety of black-box LLMs and tasks while providing intuitive and easily quantifiable evaluation results. However, consistency methods require multiple generations or sampling processes, leading to higher computational costs and complexity, as well as greater hardware resource demands and longer result calculation times. Additionally, since the query remains unchanged in each instance, if there is internal bias in the model, the confidence result may not accurately reflect correctness, and the method cannot resolve the issue of overconfidence.
\subsubsection{Self-Reflections Methods}
In black-box LLMs, self-reflection uses prompts to explore model knowledge boundaries and evaluate their own responses for reliable Confidence Estimation. Beyond this, self-reflection leverages LLMs' capability for multi-dimensional text evaluation, finding wide application in tasks like Reasoning \cite{yan2024mirror}, Translating \cite{wang2024tasteteachinglargelanguage} and Refinement \cite{madaan2024self}. While self-reflection can incorporate external feedback from diverse sources to evaluate \cite{yan2024mirror}—such as critical models \cite{peng2023check}, knowledge bases \cite{yao2022react}, or iterative self-reflection methods \cite{shinn2024reflexion}—in black-box LLMs settings, it primarily relies on the model's inherent self-reflective capabilities.

Fundamentally, self-reflection is a verbalization-based method \cite{lin2022teaching, tian2023just}. For example, \citet{lin2022teaching} designed prompts with a suffix "Confidence (0-1):" to allow the LLM to output its confidence in the original response. In practice, self-reflection prompts enable LLMs to reflect from multiple perspectives and generate confidence closer to well-calibrated values. The core of these methods lies in designing effective reflective questions and constructing feedback mechanisms. For instance, \citet{peng2023instruction} asked models to provide confidence values on a 0 to 100 scale, but experiments showed this often led to overestimated confidence. To address this, \citet{chen2024quantifying} proposed reflective questions in the form of multiple-choice items, averaged the scores, normalized them, and subsequently mapped them to corresponding confidence intervals.
% \citet{zhao2024fact} explored different prompt designs such as Chain-of-Thought (CoT)\cite{wei2022chain}, Tree-of-Thought\cite{yao2024tree}, and Self-Ask\cite{press2022measuring}, studying their impact on confidence. And The method generate intermediate confidence values, which lead to final Confidence Elicitation.

In recent methods, self-reflection not only provides confidence values but also enhances response quality. For instance, \citet{li2024think} proposed a $T^3$ prompting framework that first creates prompts to reflect on each candidate responses and generate argumentative explanations, then uses Top-K Verbalized Confidence \cite{lin2022teaching, tian2023just} to estimate the K most probable responses and their probabilities, finally averaging the confidence scores from different explanations to determine the optimal response and its confidence. \citet{zhang-etal-2024-self-contrast} proposed a framework where LLMs generate multiple diverse solving perspectives for a problem, then contrast and self-reflect on them to identify discrepancies. Based on this reflection, LLMs correct errors and produce the optimal response along with a confidence value. \citet{fang2024counterfactualdebatingpresetstances} proposed a reflective framework, CFMAD, comprising two steps: abduction generation and counterfactual debate. LLMs first assume each candidate answer is correct and generate supporting reasons. Subsequently, LLMs engage in counterfactual debates as critics and defenders, with judge LLMs evaluating the proceedings to determine the most reliable answer through confidence scoring. 

\subsubsection{Hybrid Methods}
Although both consistency and self-reflection methods can be used to obtain near-well-calibrated confidence without accessing internal parameter states, making them well-suited for black-box LLMs and widely used for reducing hallucinations, they remain post-processing techniques. As such, they cannot alter the internal parameters and states of the model and are subject to the inherent biases within the LLMs, unable to correct major logical errors. To achieve better results, recent methods combine different strategies with consistency and self-reflection techniques. For example, \citet{chen2024quantifying} proposed a dual-branch model that combines consistency and self-reflection methods. They generate multiple response samples, calculate similarity values, and design self-reflection questions to compute reflection values. These values are then normalized and combined to produce the final confidence values.
% \citet{huang2024calibrating}, on the other hand, used ChatGPT-4 to generate correctness distribution curves to quantify human assessments of response quality. They also generated confidence score distribution curves from self-consistency or self-reflection scores and then used similarity-based metrics to compute calibration errors between the curves, adjusting them to determine the optimal confidence value.

Moreover, introducing third-party models to transform black-box model into gray-box model is another common approach in Confidence Estimation for black-box LLMs. However, studies have found that directly using a white-box model, like LlaMA \cite{touvron2023llama}, to output token-level probabilities for the black-box model's response results in poor calibration due to generation differences \cite{manakul2023selfcheckgpt}. To address this issue, \citet{shrivastava2023llamas} first demonstrated that confidence values generated solely by proxy models are not always accurate. They then used confidence probabilities derived from self-consistency and verbalization methods to cross-validate with the confidence values produced by third-party proxy models, resulting in more reliable confidence estimates. \citet{ulmer2024calibrating} further attempted to train a white-box model to adapt to the generation patterns of black-box LLMs by learning the mapping between responses' confidence and correctness, thus obtaining well-calibrated confidence values. 

Additionally, many studies have introduced multiple LLMs to cross-validate and optimize responses across LLMs, ultimately combining various explanations to generate the final confidence score. For instance, \citet{zhang2023sac} utilized multiple LLMs with cross-model consistency, where responses from other LLMs for the same query are used as references to calculate the confidence value for the target black-box LLMs. \citet{feng2024don} proposed a multi-model collaboration framework, where different LLMs generate responses based on insights from their respective knowledge domains. Reflection prompts encourage these models to validate and challenge each other, reducing hallucinations and producing more reliable responses and confidence values.
% \paragraph{\textbf{Calibration Measurement Methods.}} 

\subsection{Calibration}
\subsubsection{Calibration Methods}
Calibration aims to align confidence scores with correctness. Commonly used calibration methods include temperature scaling \cite{guo2017calibration}, label smoothing \cite{szegedy2016rethinking}, and logistic calibration \cite{kull2017beta}. However, methods like label smoothing and logistic calibration are unsuitable for black-box LLMs, as they require internal model access or fine-tuning, whereas temperature scaling, which operates on the logits of model outputs, is better suited for gray-box models. In black-box LLMs, common calibration methods include post-processing or transforming the model into a gray-box. Due to the near-well-calibrated nature of Confidence Estimation in black-box LLMs, calibration pressure is reduced. This section discusses the subsequent steps required to achieve well-calibration following Confidence Estimation.

In post-processing methods, popular techniques include Histogram Binning \cite{zadrozny2001learning} and Isotonic Regression \cite{jiang2011smooth}. Histogram Binning \cite{zadrozny2001learning} adjusts confidence values within different intervals by assessing the correctness of predictions in each interval, providing a simple but potentially limited calibration when confidence distributions are uneven. Isotonic Regression \cite{jiang2011smooth}, on the other hand, seeks an optimal monotonically increasing function $f(c)$ to map the model’s confidence $ c_1, c_2, \dots, c_n $ to correctness. $f(c)$ need to satisfy the condition in Formula \ref{eq:ir}, ensuring that the confidence ranking aligns with the correctness.
\begin{equation}
\small
    f(c_1) \leq f(c_2) \leq \dots \leq f(c_n) \quad \text{for} \quad c_1 \leq c_2 \leq \dots \leq c_n
\label{eq:ir}
\end{equation}
Though flexible, Isotonic Regression is less effective in classification tasks as it doesn’t directly modify the decision boundary.

To address more complex scenarios, some methods introduce additional constraints. For instance, \citet{hebert2018multicalibration} introduced a multi-calibration method, ensuring not only overall alignment but also calibration across multiple subgroups $S_j$ through iterative alignment. The overall goal can be expressed as Formula \ref{eq:multical}. 
\begin{equation}
\small
    \min \sum_{i=1}^\mathcal{N} \sum_{S_j \subseteq \mathcal{N}} \left( \mathbb{E}[{p}_{\text{con}}(S_j)] - \mathbb{E}[p_{\text{corr}}(S_j)] \right)^2
\label{eq:multical}
\end{equation}
While this method promotes fairness, it is computationally expensive and depends heavily on subgrouping strategies, usually relying on known features. \citet{detommaso2024multicalibration} extended this to LLMs by clustering prompts in embedding space to form semi-structured groups, then applying the Improved Grouped Histogram Binning algorithm for independent group-level calibration, ensuring well-calibrated confidence both within each group and overall. This approach is better suited to generative tasks in LLMs, where decision boundaries are inherently less defined.

In addition to post-processing calibration methods, some approaches introduce proxy models to transform black-box LLMs into gray-box LLMs to facilitate calibration. For example, Bayesian calibration Network \cite{dejong1996bayesian} is a classic method that builds a probabilistic model to represent the dependency between confidence $p_{\text{con}}$ and correctness $p_{\text{corr}}$. The Bayesian Network is trained on small batches of data to estimate the conditional probability distribution $P(p_{\text{corr}} \mid p_{\text{con}})$, which is then used to infer calibrated confidence for new samples. Similarly, \citet{kuppers2021bayesian} applied a Bayesian framework with Stochastic Variational Inference to model epistemic uncertainty in calibration. By optimizing the Evidence Lower Bound, they generated calibrated confidence through a posterior distribution $q(\theta)$ of validation samples as follows Formula \ref{eq:elbo}.
\begin{equation}
\small
P(p_{\text{corr}} \mid {p}_{\text{con}}) =  \\
\frac{\mathbb{E}_{q(\theta)}[\log P({p}_{\text{con}} \mid p_{\text{corr}}, \theta)] P(p_{\text{corr}})}
{P({p}_{\text{con}})}
\label{eq:elbo}
\end{equation}

Furthermore, \citet{ulmer2024calibrating} proposed achieving calibration for each subgroup by training an auxiliary model to minimize the mean squared error between confidence and correctness. This method uses sentence embeddings (such as Sentence-BERT \cite{reimers2019sentence}) to cluster similar questions and uses the accuracy within each cluster as correctness labels. \citet{ye2021can} developed a calibration model using random forests to dynamically adjust confidence, and employed Local Interpretable Model-agnostic Explanations and SHapley Additive exPlanations to generate local explanations of model predictions, extracting meaningful features for calibration. \citet{zhao2024pareto} proposed a calibration method based on Pareto optimization, which trains a probabilistic model $h^*$(e.g., BERT \cite{devlin2018bert} or BiomedBERT \cite{gu2021domain}) to align LLMs' confidence scores with true error rates. This is achieved by integrating multiple external information sources, constructing a multi-objective loss function, and optimizing it via Pareto aggregation for effective calibration. As human perception becomes increasingly relevant as an evaluation criterion, some approaches shift calibration objectives toward human-related metrics. For instance, \citet{hashemi2024llm} trained small feedforward neural networks as calibration models using LLM and human feedback to align LLM confidence with human judgment, and fine-tuned based on feedback from different reviewers. Notably, they designed a questionnaire-based evaluation framework to assess various aspects of human perceptions regarding text quality.

\subsubsection{Measurement Methods}
In Calibration, measurement methods are needed to assess the distance between the confidence and correctness, ensuring that the confidence is well-calibrated. These methods are generally applicable, unaffected by the model's nature—whether white-box or black-box—or the specific task characteristics. Calibration measurement Methods are typically categorized into two types: \textit{Error-Based} and \textit{Correlation-Based} calculation methods. When using \textit{Error-Based} methods, correctness $y$ and confidence $p$ must be in the same measurement space to directly evaluate the model's calibration ability for specific samples, with a primary focus on detecting overconfidence. Common methods include the ECE series \cite{naeini2015obtaining, blasiok2023smooth, si2022re, naeini2015obtaining} and Brier Score \cite{brier1950verification}. The general formula is typically expressed as Formula \ref{eq:calierror}:
\begin{equation}
\text{Cal\_Error} = \sum_{i=1}^{N} w_i \cdot f(p_i, y_i)
\label{eq:calierror}
\end{equation}

In Formula \ref{eq:calierror}, $i$ represents a sample from the dataset of size $N$, and $w$ is the weight assigned to the $i$-th sample.
When using \textit{Correlation-Based} methods, correctness $Y$ and confidence $P$ do not need to follow the same scoring scale, typically evaluating overall calibration performance. For example, in \cite{zhang2024luq}, calibration error is calculated by comparing the ranking of test samples based on confidence with the ranking based on FActScore \cite{min-etal-2023-factscore} (i.e. correctness). This type of method is also known as "Relative Confidence" \cite{geng2024survey}, Common methods include AUROC \cite{boyd2013area} and AUARC \cite{nadeem2009accuracy}. The general formula is typically expressed as Formula \ref{eq:csa}:
\begin{equation}
\text{Cal\_Error} = \int_{(P, Y) \in \mathcal{D}} g(P, Y; \theta) \, d\mathcal{D}(P, Y)
\label{eq:csa}
\end{equation}.

In Formula \ref{eq:csa}, \( D \) is the joint distribution of  \( P \) and \( Y \) for dataset. \( g \) is the measurement function defined according to the specific calibration method, and controlled by the parameter \( \theta \). \( \theta \) defines the specific measurement rules of the method, such as ranking information, linear correlation, threshold, etc. The integral symbol \( \int \) represents a holistic measurement across all combinations of $(P, Y)$, providing a comprehensive evaluation of the calibration error over the dataset.

It is worth noting that in generation tasks and uncertainty estimation tasks, \textit{Correlation-Based} calculations are often used, as they help assess calibration error in complex scenarios like long-form text generation and accommodate multiple correctness scoring standards (e.g., Factuality and Coherence metrics in Summarization tasks \cite{spiess2024quality}). In classification tasks, \textit{Error-Based} calculations are more popular as they help measure calibration error for each class and identify potential knowledge gaps. We summarize various calibration measurement Methods in Table \ref{tab:cmc}, with further details available in the Appendix \ref{a1}.

\section{Applications}
\label{app}
The Calibration Process enhances both the reliability of LLMs in practical applications and the level of user trust. Based on different objectives, the major applications are categorized into two aspects: \textit{Risk Assessment and Mitigation} focusing on objective safety goals, and \textit{Human-LLM Trust Enhancement and Optimization} addressing subjective human perception goals.

\textbf{Risk Assessment and Mitigation.} Black-box LLMs face challenges in high-risk domains like medical diagnosis and autonomous driving due to potential hallucinations and overconfidence \cite{liu2023trustworthy, xiao-etal-2022-uncertainty}. The Calibration Process helps mitigate these risks, enabling practical deployments. For instance, \citet{qin2024enhancing} proposed 'Atypical Presentations Recalibration' in healthcare, while \citet{savage2024large} demonstrated improved medical diagnosis through sample consistency calibration. In production environments, \citet{zellinger2024efficiently} proposed HCMA, a framework for risk management based on uncertainty evaluations, while \citet{shen2023formalism} introduced the calibration method DwD to address compound decision risks.

\paragraph{\textbf{Human-LLM Trust Enhancement and Optimization.}} In human-machine collaboration, user trust is critical and depends on LLM response quality \cite{do2024facilitating}. Beyond factuality calibration \cite{zhao2024fact}, research focuses on aligning with user preferences. \citet{steyvers2024calibration} studied LLMs' confidence calibration based on human perception, while \citet{tsai2024efficient} developed preference-aligned decision-making processes for black-box LLMs. \citet{do2024facilitating} enhanced user-LLM trust through calibrated Factuality Scores and Source Attributions, while \citet{spiess2024calibration} demonstrated that calibration improves efficiency and accuracy in code generation tasks, a typical user-intensive application.

\section{Future Direction}
\label{fd}
\paragraph{\textbf{Developing Comprehensive Calibration Benchmarks.}} In the Calibration Process, the definition of correctness varies widely across different tasks, directly affecting the calibration objectives. For instance, some tasks focus on evaluating factual accuracy, while others prioritize dimensions like logical coherence, claim consistency, or human satisfaction. Moreover, in complex generative tasks, more granular and multi-dimensional calibration methods are often required to assess multiple aspects of the output. Therefore, developing a multi-dimensional, multi-stage benchmark that accommodates the complexity of diverse tasks is critical to improving model calibration performance.

% \textbf{General Calibration Methods for Black-Box and white-box LLMs}. Although black-box LLMs currently outperform white-box LLMs, we have observed that the iteration speed of white-box LLMs is increasing rapidly. While their overall performance has not yet surpassed black-box LLMs, White-Box models show potential in outperforming Black-Box models in certain smaller tasks. Therefore, it is crucial to develop general Confidence Estimation and Calibration Alignment methods applicable to both Black-Box and white-box LLMs.

\paragraph{\textbf{Bias Detection and Mitigation for black-box LLMs.}} Bias detection and mitigation remain critical challenges in black-box LLMs. The inaccessibility of model internals makes it difficult to identify and address bias, imposing inherent limitations on both confidence estimation and calibration. This often leads to disproportionately high or low calibration errors for specific groups or tasks, undermining the model's reliability and fairness. Traditional methods have limited effectiveness in addressing bias within black-box settings, underscoring the necessity for tailored solutions. Addressing these challenges requires a focus on improving output transparency, integrating external calibration mechanisms, and leveraging user feedback for dynamic confidence adjustment. These strategies not only tackle the unique obstacles of black-box settings but also enhance the calibration performance and practical utility of LLMs, enabling more reliable and interpretable applications across diverse real-world scenarios.
% \noindent\textbf{Balancing Cost and Calibration Effectiveness Methods}. black-box LLMs calibration methods heavily rely on input-output information. Whether using post-processing methods or proxy models, the time and cost involved are relatively high. Moreover, due to the limited access to internal information in Black-Box models, calibration results can be unstable when biases exist within the model. Thus, introducing White-Box resources to find methods that balance cost and calibration effectiveness in black-box LLMs is a meaningful research direction for the future.

% \paragraph{\textbf{Prompt Auto-Optimization.}} In black-box LLMs, prompts are crucial for estimating model information. Confidence Estimation methods in black-box LLMs, whether based on Consistency or Self-Reflection, are highly dependent on prompt design. However, the optimal prompt may vary across tasks, and even with prompt ensembles, the responses may exhibit inconsistency or redundancy, making it challenging to ensure reliable confidence estimation while keeping debugging costs low. Therefore, designing an automatic prompt optimization mechanism tailored to task-specific Confidence Estimation in black-box LLMs is essential.

\paragraph{\textbf{Calibration for Long-form Text.}} Calibrating long-form text poses significant challenges for both white-box and black-box LLMs. As the demand for LLMs grows, long texts in QA and other generative tasks often encompass multiple claims and facts, making calibration increasingly complex. Assessing whether confidence scores are well-calibrated becomes particularly difficult when only portions of the text are correct, and quality evaluation is further complicated by its inherently subjective nature. Addressing these challenges requires effective methods that incorporate human perception as a metric for correctness. Additionally, it is crucial to develop approaches tailored to black-box settings, where model parameters remain inaccessible.

% \noindent\textbf{Multimodal Calibration}. Current calibration methods are mostly designed for unimodal data, whereas multimodal data typically possess different characteristics, and confidence estimation methods may differ across modalities. As a result, unimodal methods are difficult to transfer to multimodal scenarios. Additionally, multimodal data may suffer from imbalanced distributions, leading to overconfidence or underconfidence in certain modalities. Thus, addressing calibration challenges in multimodal settings remains a significant unresolved issue. With the growing adoption of multimodal black-box LLMs in a wide range of tasks, developing calibration methods tailored to multimodal data has become a critical future research need.

\section{Conclusion}
In this survey, we comprehensively review the current research on calibration in black-box LLMs. Since the internal states of black-box LLMs are inaccessible, the available calibration methods are limited. However, due to the exceptional performance and wide application of black-box LLMs, calibration methods for these models have recently garnered significant attention. We begin by discussing how calibration can help mitigate Hallucinations and Overconfidence problems in black-box LLMs. Then, we systematically define the Calibration Process as consisting of two components: Confidence Estimation and Calibration, with the latter involving methods to measure calibration error. In the following sections, we delve into the challenges each component faces in black-box LLMs and introduce the latest research methods. Finally, we explore the applications where the Calibration Process can be beneficial in black-box LLMs and propose directions for future research. To the best of our knowledge, although numerous studies have explored calibration in black-box LLMs, no comprehensive survey has been conducted to summarize this body of work, highlighting the pioneering nature of our contribution.

\section*{Limitations}
This survey has the following key limitations:
\paragraph{\textbf{Scope Limitation.}} While our survey provides a detailed examination of Calibration Process methods applicable to black-box LLMs, it does not extensively cover approaches for white-box LLMs. As mentioned previously, the majority of existing calibration research focuses on white-box LLMs, meaning our survey may not encompass some well-known and widely recognized methods in the field. Nevertheless, conducting a survey specifically for black-box LLMs is still significant, as black-box methods are often applicable to white-box settings and may offer new perspectives for white-box LLMs Calibration Process research.

\paragraph{\textbf{Coverage Limitations}} We have made a concerted effort to comprehensively collect and summarize Calibration Process methods for black-box LLMs. However, as the Calibration Process often appears in many studies as an auxiliary optimization or evaluation tool rather than a central focus, some important works on calibration processes might have been overlooked.

\section*{Ethics Statement}
This study involves reviewing published academic literature on Calibration Process for black-box LLMs. As it does not involve human subjects, experimental procedures, or sensitive datasets, there are no ethical concerns or risks. The research follows ethical guidelines for academic integrity, and the researchers declare no conflicts of interest. All data collection adheres to standard ethical principles for systematic literature reviews.

% \section*{Acknowledgments}

\bibliography{acl2023}
\bibliographystyle{acl_natbib}

\clearpage

\appendix

\section{Appendix}
\label{sec:appendix}

\begin{table*}[h]
\scriptsize
% \small
\begin{tabularx}{\linewidth}{m{3cm}<{\centering} m{4cm}<{\centering} m{5cm}<{\centering} m{2cm}<{\centering}}
	\hline
Measurement Methods & Measurement Focus & Preferred Task & Category \\ 
 
    \hline
ECE \cite{naeini2015obtaining}, MCE \cite{naeini2015obtaining}, MacroCE \cite{si2022re}, TL-ECE \cite{gupta2021top}, smECE \cite{blasiok2023smooth}, Reliability Diagram \cite{arrieta2022metrics} & Measures the difference between confidence and correctness via interval division; Reliability Diagram for visualization. & Sentiment Analysis. Natural Language Inference. Multiple-choice QA. Truthful QA. Web QA. Reasoning Tasks & Error-Based methods \\
    \hline
Brier Score \cite{brier1950verification}, BSS \cite{bradley2008sampling}, Quadratic Score \cite{selten1998axiomatic}, Logarithmic Score \cite{bickel2007some}, NLL \cite{gneiting2007strictly} & Measures the overall difference between model confidence and true labels. & Sentiment Analysis. Natural Language Inference. Multiple-choice QA. & Error-Based methods  \\
    \hline
AUROC \cite{boyd2013area}, AUPRC \cite{qi2021stochastic}, Weighted AUROC \cite{keilwagen2014area}, AUARC \cite{nadeem2009accuracy} & Measures the model's ability to distinguish correct from incorrect at different confidence thresholds. & Reasoning Tasks: Arithmetic, Symbolic, Professional Knowledge, Ethical, Commonsense. Summarization. Hallucination Detection. Open-ended QA. Conversational QA. Natural Language Generation. Truthful QA. & Correlation-Based methods   \\
    \hline
PCC \cite{cohen2009pearson}, SCC \cite{sedgwick2014spearman}, RCE \cite{huang2024uncertainty} & Measures the correlation or ranking consistency between model confidence scores and correctness. & Hallucination Detection. Multiple-choice QA. Long-form text QA. Commonsense QA. RiddleSense Reasoning& Correlation-Based methods \\
\hline
\end{tabularx}
    \caption{Table of calibration measurement Methods. ECE=Expected Calibration Error, MCE=Maximum Calibration Error, MacroCE=Macro Calibration Error, TL-ECE=Top Label Expected Calibration Error, smECE=Smooth Expected Calibration Error, BSS=Brier Skill Score, NLL=Negative Log-Likelihood, AUROC=Area Under the Receiver Operating Characteristic Curve, AUPRC=Area Under Precision-Recall Curve, AUARC=Area Under the Accuracy-Rejection Curve, PCC=Pearson Correlation Coefficient, SCC=Spearman's Rank Correlation Coefficient, RCE=Rank-Calibration Error}
    \label{tab:cmc}
\end{table*}

\subsection{Calibration Measurement Methods}
\label{a1}
This section provides a comprehensive explanation of the Error-Based Methods and Correlation-Based Methods within calibration measurement Methods. Additionally, these methods are summarized in Table \ref{tab:cmc}.
\subsubsection{Error-Based Methods}
In Error-Based calibration measurement Methods, the ECE (Expected Calibration Error) \cite{naeini2015obtaining} series of methods is most widely adopted. In the ECE \cite{naeini2015obtaining}, the weight coefficient \(w_i\) in the general error calculation Formula \ref{eq:calierror} can be defined based on the number of test samples or confidence interval weights, typically expressed as \(\frac{|B_b|}{N}\), where \(B_b\) represents the $b$-th confidence interval, and \(|B_b|\) denotes the number of samples within that interval. The error term \(f(p_i,y_i)\) quantifies the deviation between predicted confidence probabilities and actual correctness probabilities, typically expressed as \(\left|\frac{1}{|B_b|}\sum_{i\in B_b}p_i-\frac{1}{|B_b|}\sum_{i\in B_b}y_i\right|\).
Furthermore, both Brier Score \cite{brier1950verification} and Negative Log-Likelihood (NLL) \cite{gneiting2007strictly} are commonly employed. In the Brier Score, all test samples are assigned equal weights, with weight coefficient \(w_i=\frac{1}{N}\), and its error term \(f(p_i,y_i)\) utilizes the squared difference between predicted confidence probabilities and actual correctness probabilities, expressed as \((p_i-y_i)^2\). Similarly, the NLL method employs uniform weight distribution but defines its error term \(f(p_i,y_i)\) in cross-entropy form: \(-\left(y_i\log(p_i)+(1-y_i)\log(1-p_i)\right)\).

\subsubsection{Correlation-Based Methods}
In Correlation-Based calibration measurement Methods, both Spearman's Rank Correlation Coefficient \cite{sedgwick2014spearman} and Pearson Correlation Coefficient \cite{cohen2009pearson} are used to assess the relationship between confidence scores and correctness. Spearman focuses on ranking consistency, where the parameter $\theta$ in the general Formula \ref{eq:csa} represents the ranking information $r(C)$ and $r(Y)$ of confidence and correctness within the distribution $\mathcal{D}$. For the $i$-th sample, ranks $r(C_i)$ and $r(Y_i)$ are assigned based on its confidence score $C_i$ and correctness score $Y_i$, respectively. The measurement function $g$ is defined as follows:
\begin{equation}
\small
g_{\text{Spearman}}(C, Y; \theta) = 1 - \frac{6 \sum_{i=1}^{N} (r(C_i) - r(Y_i))^2}{N(N^2 - 1)}
\label{eq:spear}
\end{equation}

Thus, Spearman's calibration error evaluates the overall ranking deviation within the distribution $\mathcal{D}$. On the other hand, Pearson measures the linear relationship between confidence scores and correctness, where $\theta$ in the general Formula \ref{eq:csa} represents the mean parameters $\bar{C}$ and correctness $\bar{Y}$ associated with linear correlation. The measurement function $g$ is defined as:
\begin{equation}
\small
g_{\text{Pearson}}(C, Y; \theta) = \frac{\sum_{i=1}^{N} (C_i - \bar{C})(Y_i - \bar{Y})}{\sqrt{\sum_{i=1}^{N} (C_i - \bar{C})^2} \sqrt{\sum_{i=1}^{N} (Y_i - \bar{Y})^2}}
\label{eq:pearson}
\end{equation}
Thus, Pearson’s calibration error evaluates the linear correlation between confidence and correctness scores within the distribution $\mathcal{D}$.

In addition to the above, AUROC \cite{boyd2013area} and AUARC \cite{nadeem2009accuracy} are also commonly used in Correlation-Based Methods to measure calibration errors. These methods focus on evaluating the model's ability to distinguish between correct and incorrect samples at various confidence thresholds, but their emphases differ. AUROC emphasizes the global ability of the model to separate correct and incorrect categories, reflecting overall uncertainty \cite{kuhn2023semantic}. In contrast, AUARC evaluates the trade-off between recall and precision, particularly in high-recall scenarios.
For AUROC, the parameter $\theta$ represents the labels and comparative information between correct and incorrect samples. The measurement function $g$ is defined as:
\begin{equation}
\small
g_{\text{AUROC}}(C, Y; \theta) = \frac{1}{N_c N_i} \sum_{i=1}^{N_c} \sum_{j=1}^{N_i} \mathbb{I}(C_i > C_j)
\label{eq:auroc}
\end{equation}
Here, $N_c$ and $N_i$ denote the number of correct and incorrect samples, respectively. The indicator function $\mathbb{I}(C_i > C_j)$ equals 1 when the confidence of a correct sample $C_i$ exceeds that of an incorrect sample $C_j$, and 0 otherwise. Thus, AUROC evaluates the model’s global ability to distinguish correct and incorrect samples within the distribution $\mathcal{D}$.
For AUARC, the parameter $\theta$ represents the confidence threshold $\tau$. Its measurement function $g$ is expressed as:
\begin{equation}
\small
g_{\text{AUARC}}(C, Y; \theta) = \frac{1}{N} \sum_{i=1}^{N} \mathbb{I}(C_i \geq \tau) \cdot A(C_i, Y_i)
\label{eq:auarc}
\end{equation}
Here, $\tau$ is the confidence threshold, and $\mathbb{I}(C_i \geq \tau)$ is an indicator function that equals 1 if the confidence score $C_i$ of the $i$-th sample exceeds or equals $\tau$, indicating acceptance of the sample; otherwise, it equals 0, rejecting the sample. $A(C_i, Y_i)$ represents the correctness of accepted samples. Therefore, AUARC evaluates the overall correctness of accepted samples at different confidence thresholds $\tau$. For well-calibrated models, the overall correctness improves when only high-confidence samples are accepted.
\end{document}